\documentclass{article}
\usepackage{spconf,amsmath,graphicx}
\usepackage{algorithmic}
\usepackage{float}
\usepackage{multirow}
\usepackage{stfloats}
\usepackage{textcomp}
\usepackage{threeparttable}
\usepackage{booktabs}
\usepackage{xcolor}
\usepackage{caption} 
\usepackage{subcaption}
 \usepackage{amssymb}

\title{Attention Based Relation Network for Facial Action Units Recognition}
\name{Yao Wei \qquad Haoxiang Wang$^{\star}$ \qquad Mingze Sun\qquad Jiawang Liu\thanks{$^{\star}$ Corresponding author: Haoxiang Wang (hxwang@scut.edu.cn).}
\thanks{This work was supported by Guangdong Basic and Applied Basic Research Foundation (2021A1515011852).}}
\address{South China University of Technology, China}
\begin{document}
%
\maketitle
\begin{abstract}
Facial action unit (AU) recognition is essential to facial expression analysis. Since there are highly positive or negative correlations between AUs, some existing AU recognition works have focused on modeling AU relations. However, previous relationship-based approaches typically embed predefined rules into their models and ignore the impact of various AU relations in different crowds. In this paper, we propose a novel Attention Based Relation Network (ABRNet) for AU recognition, which can automatically capture AU relations without unnecessary or even disturbing predefined rules. ABRNet uses several relation learning layers to automatically capture different AU relations. The learned AU relation features are then fed into a self-attention fusion module, which aims to refine individual AU features with attention weights to enhance the feature robustness. Furthermore, we propose an AU relation dropout strategy and AU relation loss (AUR-Loss) to better model AU relations, which can further improve AU recognition. Extensive experiments show that our approach achieves state-of-the-art performance on the DISFA and DISFA+ datasets.
\end{abstract}
\section{Introduction}
\label{sec:intro}

As AUs occur in different regions of the face, most current works \cite{5,6,7} treat AU recognition as a multi-label classification problem and consider AUs to be independent of each other. However, the AUs are highly related to each other. Due to the anatomical mechanisms of faces, a facial expression is associated with a certain set of AUs. For example, AU6 (Cheek Raiser) and AU12 (Lip Corner Puller) tend to be activated together when we express happy emotions, and it is difficult to make AU9 (Brow Lowerer) without the presence of AU4(Inner Brow Raiser)\cite{9}. On the other hand, some AUs are not likely to appear simultaneously because of the structural limitations controlled by facial anatomy. For example, we can hardly simultaneously make AU22(Lip Funneler) and AU23(Lip Tightener).
\begin{figure}
	\centering
		\includegraphics[scale = 0.42]{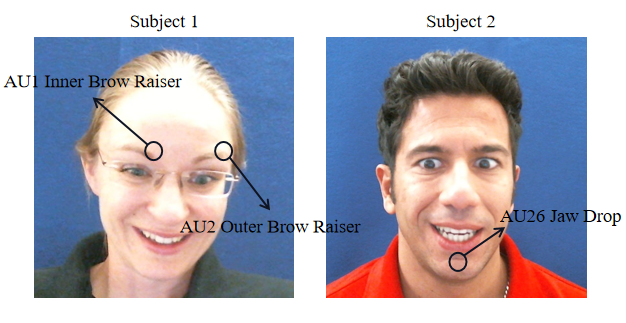}
        \caption{Illustration of various AU relations. \textit{Subject 1} tends to raise her brows (AU1 and AU2) when smiling, but \textit{Subject 2} tends to drop his jaw (AU26) when smiling. }
        \label{Pic:aurelaton}
\end{figure}

Considering the intuitive AU relations, some works have made progress in modeling the AU relationships. Robert Walecki et al. \cite{10} propose to combine deep learning with conditional random field (CRF) to model AU dependencies for more accurate AU detection. Corneanu et al. \cite{11} develop a complex model to exploit AU correlations via probabilistic graphic approaches. However, these widely adopted strategies \cite{10,11,12} do not explicitly consider the AU relationship in their model design and bring in noise information from non-AU regions, which limits the performance of AU-level relation modeling. To better model AU-level relations, Li et al. \cite{13} recently propose to apply Gated Graph Neural Network (GGNN) to learn relationship-embed AU feature representation with a defined AU relation graph based on statistics of the training data. However, due to the predefined rules, these constrained models \cite{13,14} can only learn limited AU relationships. Besides, in different crowds, people may have different AU relations. As shown in Fig.\ref{Pic:aurelaton}, some people tend to raise their brows (AU1 and AU2) when smiling, while others may tend to drop their jaw (AU26). However, existing AU relation-based methods do not consider these various AU relations in different crowds.

Driven by this observation and inspired by attention mechanism, we propose a novel ABRNet for facial AU recognition. Our main contributions are listed as follows:
\begin{itemize}
\item ABRNet is proposed to capture the various AU relations in different crowds. The ABRNet uses a relation learning module to automatically capture different AU relations and a self-attention fusion module to refine the AU features with attention weights; 
\item AU relation dropout and AU relation loss (AUR-Loss) are introduced to better utilize AU relations. It can further improve AU recognition without additional computation costs;
\item Extensive experiments show that our model achieves state-of-the-art performance on two benchmark datasets: DISFA and DISFA+.
\end{itemize}

\begin{figure*}[ht]
\centering{\includegraphics[scale=0.45]{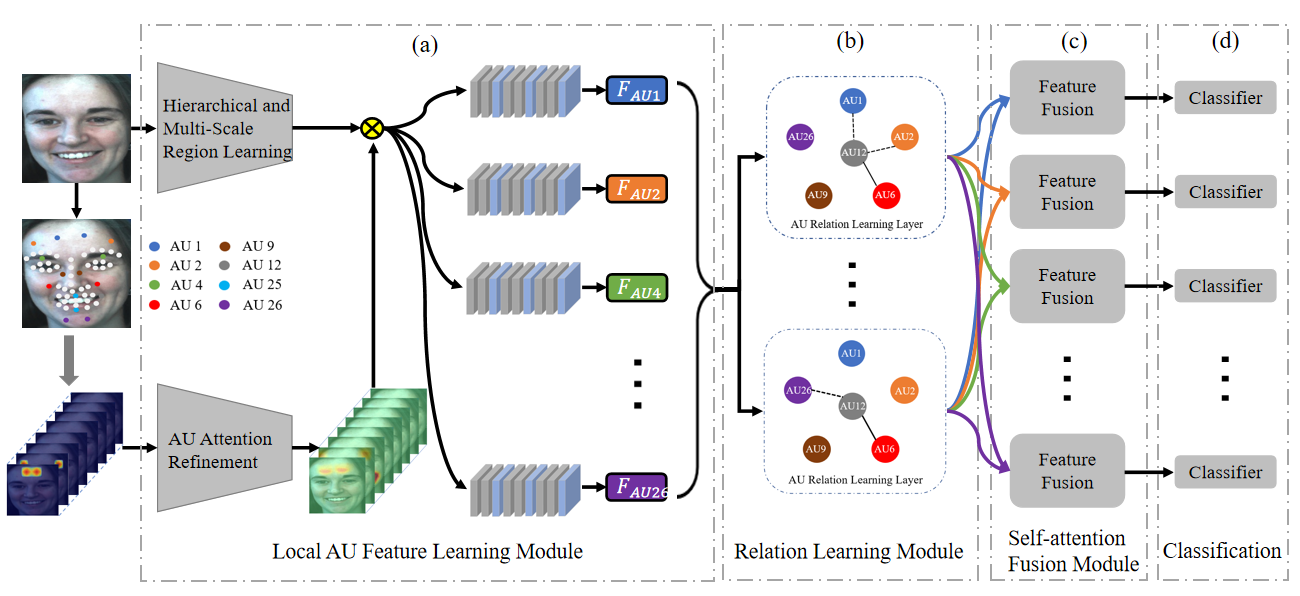}}
\caption{The overall architecture of the proposed ABRNet.}

\label{Picnet}
\end{figure*}
\section{Proposed method}

The pipeline of our ABRNet is illustrated in  Fig.\ref{Picnet}. It mainly consists of a local AU feature module(detailed in Section \ref{local au feature learning}), a relationship learning module (detailed in Section \ref{relation_modile}), and a self-attention fusion module (detailed in Section \ref{fusion_module}). An AUR-Loss and an AU relation dropout strategy are proposed and adopted for training optimizations, which are introduced in Section \ref{training optimization}.

\subsection{Local AU Feature Learning}
\label{local au feature learning}
Given a face image \emph{I}, the hierarchical and multi-scale region learning module is used to extract multi-scale AU representations from each local patch. We predefine attention maps according to the facial landmarks and utilize the AU attention refinement module to get the refined attention maps \emph{A} =\{$A_1, A_2, ..., A_n$\}, where \emph{n} is the number of AUs. The structure of hierarchical and multi-scale region learning module and AU attention refinement module is referred to \cite{8}. The multi-scale feature is element-wise multiplied with the refined attention map of each AU to obtain AU feature. Then, each branch is performed with a network consisting of three convolution blocks, each of which consisting two plain convolutional layers and a max-pooling layer. The last max-pooling layer is followed by a fully-connected layers with the dimensions of $d_l$ to learn local AU feature \emph{F} =\{$F_1, F_2, ..., F_n$\}.

\subsection{Relation Learning Module}
\label{relation_modile}
The key component in the relation learning layer is the calculation of $R_{i}^l$, $i.e.$, the feature of \emph{i}-th AU learned in \emph{l}-th relation learning layer, which involves relation coefficient and AU dictionary. AU dictionary \emph{F}=\{$F_1, F_2, ..., F_n$\} is extracted from the local AU feature learning module. The relation coefficient $\mu_{ij}^l$ represents the interaction of \emph{j}-th AU on \emph{i}-th AU, which is computed using Scaled DotProduct Attention \cite{attention} as shown in Eq. \ref{relation_coefficient}.

\begin{equation}
\centering
\label{relation_coefficient}
\mu_{ij}^l=\operatorname{softmax}\left(\sigma \left(\frac{\left(W_{\mathrm{Q}} F_{i}\right)^{T}\left(W_{\mathrm{K}} F_{j}\right)}{\sqrt{d_{\mathrm{m}}}}\right)\right)
\end{equation}
where $\sigma$ denotes the activation Leaky ReLU function, $W_{\mathrm{Q}} \in \mathbb{R}^{d_{\mathrm{m}} \times d_{\mathrm{l}}}$ and $W_{\mathrm{K}} \in \mathbb{R}^{d_{\mathrm{m}} \times d_{\mathrm{l}}}$ are weight parameters learned. 

To learn the highly correlated AU relationships, we weigh the aggregation of top-\emph{k} AU features to retain only \emph{k} most important relational information of each AU, as shown in Eq. \ref{top_k}.
\begin{equation}
\label{top_k}
R_{i}^l=\sigma \left(\sum_{j \in \mathcal{A}_{i}^{k}} \mu_{ij}^lF_j \right)
\end{equation}
where the relation coefficient $u_{ij}^l$ is used to aggregate relational messages from AU dictionary to generate output feature for each AU, and $\mathcal{A}_{i}^{k}$ denotes a set consisting of AUs with \emph{k} largest relation coefficient to \emph{i}-th AU. 

After each AU relation learning layer, we get a set of refined AU features $R^l$ =\{$R_{1}^l, R_{2}^l, ..., R_{n}^l$\}, which are expected to have all the useful relational information from other AUs. We assume that there are various AU relations in different crowds. Hence, multiple relation learning layers are put in parallel as our relation learning module, which can model different AU relations and learn the features $R$  =\{$R^1, R^2, ..., R^m$\}, where \emph{m} denotes the number of AU relation learning layers.

\subsection{Self-attention Fusion Module}
\label{fusion_module}
Considering that an AU may have a different status under different AU relations, we design a self-attentive fusion module to fuse AU features from different relation learning layers with attention weights rather than simply concatenating them. The fusion module employs a FC layer and a sigmoid activation function to estimate attention weights. The attention weight of the \emph{i}-th AU in \emph{l}-th relation learning layer is formulated by: $\beta_{i}^l$= $f\left(R_{i}^lq^0 \right)$, where $q^0$ is the parameter of FC, $f$ denotes the sigmoid function. In this stage, we summarize all the AU features from different relation learning layers with their attention weights into a global AU representation, which is formulated as:
\begin{equation}
\label{au_repre}
S_i= \sum_{l=0}^{m} \beta_{i}^l R_{i}^l
\end{equation}
where $S_i$ is a compact representation of \emph{i}-th AU. Finally, the AU representations, \emph{i.e.}, $S$ =\{$S_{1}, S_{2}, ..., S_{n}$\}, are fed to a multi-classifier to obtain the predicted occurrence probability $\hat{p}_i$ of each AU.

\subsection{Training Optimizations}
\label{training optimization}
For better utilizing AU relations, AU relation dropout strategy and AUR-Loss are proposed as training optimizations.

The Fusion module can assign different attention to various AU relations. However, the different relation learning layers may learn the same AU relation rather than various AU relations. To solve this problem, we propose an AU relation dropout strategy. We randomly remove (\emph{m} - \emph{t}) relation learning layers and use a subset of \emph{t} relation learning layer to compute the global AU representation, $i.e. R_i$, to train each sample:
\begin{equation}
\label{dropout}
S_i= \frac{m}{t} \sum_{l \in \mathcal{O}_{i}^{t}} \beta_{i}^l R_{i}^l
\end{equation}
where $\mathcal{O}_{i}^{t}$ denotes a set consisting \emph{t} randomly selected relation learning layers. For inference, we consider the output features of all the relation learning layers, as in Eq. \ref{au_repre}.

The weighted multi-label cross-entropy loss is adopted for multi-label binary classification task, computed as:
\begin{equation}
\label{cross-entropy}
\mathcal{L}_{\text {cross}}= - \sum_{i=1}^{n} w_{i}\left[p_{i} \log \hat{p}_{i}+\left(1-p_{i}\right) \log \left(1-\hat{p}_{i}\right)\right]
\end{equation}
where $p_i$ denotes the ground-truth probability of occurrence for the \emph{i}-th AU, which is 1 for occurrence and 0 otherwise. $w_i$ is the data balance weights employed in \cite{8}, computed as:$w_i$ = $1/r_i/\sum_{u=1}^{n}(1/r_u)$,where $r_i$ is the occurrence rate of the i-th AU in the training set.

In many cases, AUs appear in combination rather than separately. However, the cross entropy loss in Eq.\ref{cross-entropy} makes the AU prediction independent. In order to make the model focus more on the prevalent combination of AUs (\emph{i.e.}, the universal relationship between AUs) in the training, we design an AU relation loss (AUR-Loss) based on the relationship among individual AUs. The AU relationships are calculated using the conditional probability on the training samples, which is formulated as:

\begin{equation}
\label{pos_relation}
r_{ij}=P\left(y_{j}=1 \mid y_{i}=1\right)-P\left(y_{j}=1\right)
\end{equation}
\begin{equation}
\label{relation}
a_{ij}=\left\{\begin{array}{ll}P\left(y_{j}=1 \mid y_{i}=1\right), \text{ if } r_{ij} \textgreater 0;
\\
\frac{\displaystyle P\left(y_{j}=1 \mid y_{i}=1\right)}{\displaystyle P\left(y_{j}=1\right)}-1,\text{ if } r_{ij}\leq 0;
\end{array}\right.
\end{equation}
where $y_n$ denotes the \emph{n}-th AU label. A positive/negative value of $a_{ij}$ represents a positive/negative relationship between $AU_i$ and $AU_j$, and $\left| a \right|$ represents the strength of the relationship. The AUR-Loss is formulated as:

\begin{equation}
\label{AUR_Loss}
\mathcal{L}_{\text {AUR}}=\sum_{i=1}^{n} \sum_{j=1}^{n} w_{i}\mathcal{L}_{AU_{ij}}
\end{equation}

In Eq. \ref{AUR_Loss}, $\mathcal{L}_{AU_{ij}}$ is defined as:
\begin{equation}
\label{Loss_ij}
\begin{split}
\mathcal{L}_{AU_{ij}} = \left\{\begin{array}{ll}
max\left[2*a_{ij}-\hat{p}_{i}-\hat{p}_{j}, 0\right], 
\\\text { \quad if } a_{ij}>p_{pos}, p_i=1, p_j=1;
\\ max\left[-a_{ij}-\hat{p}_{i}+\hat{p}_{j}, 0\right],
\\\text { \quad  if } a_{ij}<p_{neg}, p_i=1, p_j=0; 
\\ 0,\text{ if } p_{neg} \leq  a_{ij} \leq p_{pos} \text{ or } p_i=0 ;
\end{array}\right.
\end{split}
\end{equation}
where $p_{p}$ and $p_{n}$ is the threshold for positive and negative relationship, respectively. Finally, the joint loss of our ABRNet is defined as:
\begin{equation}
\label{loss_all}
\mathcal{L} =\mathcal{L}_{\text {cross}}+\lambda \mathcal{L}_{\text {AUR}}
\end{equation}
where $\lambda$ is a trade-off parameter.

\section{EXPERIMENTS}
\subsection{Experimental Setup}
We evaluate our method on two popular AU recognition datasets, \emph{i.e.} DISFA \cite{disfa} and DISFA+\cite{disfa_plus}. To be consistent with comparative models, the F1-scores of 8 selected AUs are reported on three-fold subject-independent cross-validation for all experiments.

All the face images are aligned and resized to 200 × 200. During training, the input image is randomly cropped to the size of 176 × 176 and processed with data enhancement, including random horizontal flip and random scaling. The facial landmarks are detected using \cite{landmark_detector}. The parameters with respect to the structure of ABRNet are chosen as $d_l$ = 512 and \emph{m} = 4. Besides, we set $d_m$ = 256 (described in Eq. \ref{relation_coefficient}), $\lambda$ = 0.1, \emph{t} = 2, $p_{p}$ = 0.5 and $p_{n}$ = -0.7. The classifier is composed of two fully-connected layers with the dimensions of 64 and 2. Our ABRNet is trained with the stochastic gradient descent (SGD) solver, a Nesterov momentum of 0.9 and a weight decay of 0.0005. Learning rate is set to 0.01 initially, and multiplied by a factor of 0.5 every 2 epochs.

\subsection{Different Experimental Settings}
This subsection investigates the different settings of ABRNet, including different number of relation learning layers ($i.e.$ \textit{m}), different fusion schemes, different AU relation dropout ratio ($i.e.$ 1-\emph{t}/\emph{m}), and different parameters ($i.e.$ $\lambda$) in AUR-Loss. The experimental results are displayed in Table \ref{tab:array}.

Increasing \textit{m} from 0 to 4 gradually improves the performance, while larger \textit{m} (\textit{m} $>$ 4) may degrade the performance, which indicates that a stack of 4 relation learning layers is sufficient to capture the different AU relations from different crowds. There is no significant improvement in concatenating features compared to average pooling. The score average slightly improves the average pooling by 0.5\% while our self-attention fusion (indicated by S-F) outperforms the average pooling by 1.2\%. The AU dropout ratio of 0.5 performs better than the others. AU dropout strategy can help the relation learning layers learn different AU relations and increase the F1-score by 0.8\%. The performance peaks at $\lambda$ = 0.1. Larger $\lambda$ ($i.e.$ 0.15 and 0.2) degrades the performance, which can be explained that original AU features might be ignored if too much attention is paid to AU relations.

\begin{table}
    \centering
    \caption{Comparison of different settings on DISFA+.}
    \begin{subtable}[t]{0.49\linewidth}
    \centering
    \caption{Different \textit{m}}
        \begin{tabular}{cc}
            \toprule
            m & F1-score \\
            \midrule
            0 & 78.3 \\
            2 & 80.2 \\
            4 & \textbf{81.1} \\
            6 & 80.3\\
            \bottomrule
        \end{tabular}
    \end{subtable}
    \begin{subtable}[t]{0.49\linewidth}
    \caption{Different fusion schemes}
    \centering
        \begin{tabular}{ccc}
            \toprule
            Fusion Scheme & F1-score \\
            \midrule
           Average pooling & 79.9\\
           Feat. Concat. &80.0\\
           Score avg. & 80.4\\
           S-F(ours) &  \textbf{81.1}\\
            \bottomrule
        \end{tabular}
    \end{subtable}

    \quad
    
    \begin{subtable}[t]{0.49\linewidth}
    \caption{AU dropout ratio}
    \centering
        \begin{tabular}{cc}
            \toprule
            1-\emph{t}/\emph{m} & F1-score \\
            \midrule
            0 & 80.3\\ 
            0.25 & 80.8\\
            0.5 &  \textbf{81.1}\\
            0.75 &  79.7 \\
            \bottomrule
        \end{tabular}
    \end{subtable}
    \begin{subtable}[t]{0.49\linewidth}
    \caption{Different $\lambda$}
    \centering
        \begin{tabular}{ccc}
            \toprule
            $\lambda$ & F1-score  \\
            \midrule
            0 &  80.3\\
            0.05 &  80.8\\
            0.1 &  81.1\\
            0.2 &  80.4\\
            \bottomrule
        \end{tabular}
    \end{subtable}
    \label{tab:array}
\end{table}

\begin{table}[t]
\centering
\caption{Comparison of F1-score (\%) on DISFA.}
    \begin{threeparttable} 
        \begin{tabular}{ c| c c c c c }
        \toprule[1.1pt]
        AU & ARL& UGN-B& J\^{A}A & MONET& ABRNet\\
        \midrule[0.7pt]
        1 & 43.9 & 43.3 & 62.4 &55.8 &\textbf{68.1}\\
        2 & 42.1 & 48.1 & 60.7 &60.4 &\textbf{68.6}\\
        4 & 63.6 & 63.4 & 67.1 &68.1 &\textbf{72.9}\\
        6 & 41.8 & 49.5 & 41.1 &\textbf{49.8} &37.1\\
        9 & 40.0 & \textbf{48.2} & 45.1 &48.0 &39.2\\
        12 & 76.2 & 72.9 & 73.5 &73.7 &\textbf{78.7}\\
        25 & 95.2 & 90.8 & 90.9 &92.3 &\textbf{93.6}\\
        26 & 66.8 & 59.0 & \textbf{67.4} &63.1 &59.3\\
        \midrule[0.7pt]
        Avg & 58.7 & 59.4 & 63.5 &63.9 &\textbf{64.7}\\
        \bottomrule[1.1pt]
        \end{tabular}
\end{threeparttable}  
\label{tab:disfa}
\end{table}

\begin{table}[t]
\centering
\caption{Comparison of F1-score (\%) on DISFA+.}
    \begin{threeparttable} 
        \begin{tabular}{ c| c c c  c }
        \toprule[1.1pt]
        AU & DRML& AU R-CNN& J\^{A}A& ABRNet\\
        \midrule[0.7pt]
        1 & 29.1& 48.7 & \textbf{80.7}&76.4\\
        2 & 24.6& 43.6 & \textbf{81.6}&72.3\\
        4 & 53.2& 57.5 & 79.5&\textbf{83.5}\\
        6 & 51.2& 47.2 & \textbf{78.4}&77.2\\
        9 & 35.3& 33.2 & 68.8&\textbf{83.2}\\
        12 & 37.6& 48.6 & 85.4&\textbf{87.6}\\
        25 & 55.6& 63.4 & 88.3&\textbf{94.6}\\
        26 & 43.4& 36.8 &71.2 &\textbf{74.3}\\
        \midrule[0.7pt]
        Avg & 41.3 & 47.4 & 79.2 &\textbf{81.1}\\
        \bottomrule[1.1pt]
        \end{tabular}
\end{threeparttable}  
\label{tab:disfa+}
\end{table}

\subsection{Comparison with Other State-of-the-art Methods}
We compare our ABRNet against the recent state-of-the-art methods including ARL \cite{20}, UGN-B \cite{song2021}, J\^{A}A-Net \cite{8}, and MONET \cite{21} on the DISFA dataset. We re-implement DRML \cite{5}, AU R-CNN \cite{22} and J\^{A}A-Net \cite{8} to conduct comparative experiments on DISFA+ dataset.
Table \ref{tab:disfa} shows the comparison results on DISFA dataset. It can be observed that ABRNet obtains the best results in categories of AU1, AU2, AU4, AU12 and AU25. ABRNets achieves the highest average F1-score of 64.7\% for 8 AUs, which outperforms MONET (with the second-highst F1-score) by 0.8\%.AU recognition results by different methods on DISFA+ datase are shown in Table \ref{tab:disfa+}. ABRNet outperforms J\^{A}A-Net by a large margin with the highest average F1-score 81.1\% and also achieves significant outperforms for each individual AU.

\section{conclusion}
In this paper, we propose a novel attention-based relation network (ABRNet) for facial action unit recognition. ABRNet can capture the various AU relations in different crowds without the constraints of processing order or predefined rules and refine AU features with different attention weights. Furthermore, two training optimizations: AU relation dropout strategy and AUR-Loss are introduced, which can further improve AU relation learning. The proposed ABRNet is compared with recent state-of-the-art works on DISFA and DISFA+ datasets and achieves outstanding performance.

\vfill\pagebreak

\bibliographystyle{IEEEbib}
\bibliography{strings,refs}

\end{document}